\documentclass[letterpaper]{article} 
\usepackage{aaai2026}  
\usepackage{times}  
\usepackage{helvet}  
\usepackage{courier}  
\usepackage[hyphens]{url}  
\usepackage{graphicx} 
\usepackage{amsmath}
\usepackage{natbib}  
\usepackage{caption} 
\usepackage{algorithm}
\usepackage{algorithmic}

\usepackage{newfloat}
\usepackage{listings}
\DeclareCaptionStyle{ruled}{labelfont=normalfont,labelsep=colon,strut=off} 
\floatstyle{ruled}
\newfloat{listing}{tb}{lst}{}
\floatname{listing}{Listing}
\title{Towards aligned body representations in vision models}

\newcommand{\equalcontrib}{\textsuperscript{\rm \dagger}}

\author{
    Andrey Gizdov\textsuperscript{\rm1,2}\equalcontrib$^\dagger$,
    Andrea Procopio\textsuperscript{\rm 1,3}\equalcontrib,
    Yichen Li\textsuperscript{\rm 1},
    Daniel Harari\textsuperscript{\rm 2},
    Tomer Ullman\textsuperscript{\rm 1}
}
    
\affiliations{
    \textsuperscript{\rm 1}Harvard University \quad \textsuperscript{\rm 2}Weizmann Institute of Science \quad \textsuperscript{\rm 3}Bocconi University\\ \{andreygizdov, aprocopio, yichenli, tullman\}@fas.harvard.edu,\; hararid@weizmann.ac.il
}

\begin{document}
\maketitle

\begin{abstract}
Human physical reasoning relies on internal “body” representations — coarse, volumetric approximations that capture an object’s extent and support intuitive predictions about motion and physics. While psychophysical evidence suggests humans use such coarse representations, their internal structure remains largely unknown. Here we test whether vision models trained for segmentation develop comparable representations. We adapt a psychophysical experiment conducted with 50 human participants to a semantic segmentation task and test a family of seven segmentation networks, varying in size. We find that smaller models naturally form human-like coarse body representations, whereas larger models tend toward overly detailed, fine-grain encodings. Our results demonstrate that coarse representations can emerge under limited computational resources, and that machine representations can provide a scalable path toward understanding the structure of physical reasoning in the brain.
\end{abstract}

\section{Introduction}
Human perception is concerned with \emph{what} entities are present, \emph{where} the entities are, and \emph{how} a scene will unfold \cite{marr2010vision, freyd1987dynamic}. The problems of `what', `where, and `how' are coupled but separate, and it is likely that they are supported by different computations and representations. In particular, for the purposes of inferring the identity of objects (e.g. telling apart a thermos and a water bottle), it may be important to have detailed object segmentations. But for the purposes of estimating \textit{how} a physical scene will unfold, which objects will collide and where they will end up, it is sufficient and more cost-effective to have more coarse object-representations (e.g. if all that matters is catching them, a bottle and a cup can roughly be approximated by a cylinder). \\

\begin{figure}[t]
  \centering
  \makebox[0.95\linewidth]{\ttfamily\footnotesize Shape\hfill Body}

  \setlength{\tabcolsep}{0pt}
  \begin{tabular}{c}
    \includegraphics[width=0.95\linewidth]{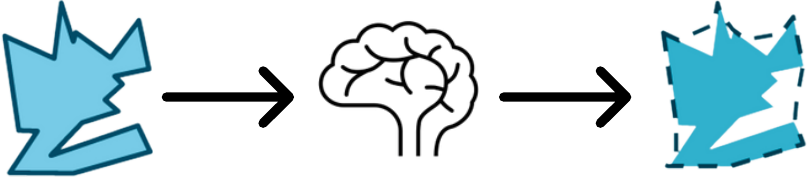} \\
    {\small (a) Humans.} \\[4pt]
    \includegraphics[width=0.95\linewidth]{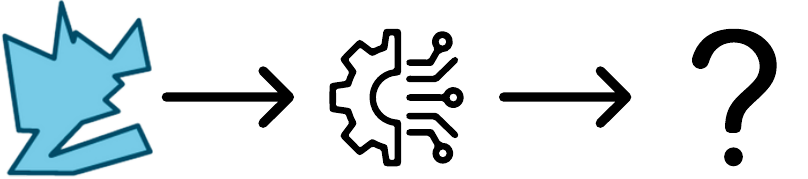} \\
    {\small (b) Vision models.}
  \end{tabular}

  \caption{Stimuli vs Body representations (dotted lines) in (A) humans and (B) vision models.}
  \label{fig:body_representation}
\end{figure}

\noindent The separation between coarse and fine-grain object segmentations for different purposes is widely used in engineering, in particular, in simulated environments such as games and animations. In humans, it has been suggested that the brain has a similar processing split between physics and graphics \citep{ullman2017mind, balaban2025physics}, with fine-grain meshes being used for the purposes of rendering and recognition, and coarse-grain bodies for the purposes of prediction and action. This split maps onto a neural division between the dorsal and ventral streams in human vision, and there is psychophysical evidence that humans make use of coarse body approximations \cite{li2023approximate} (see Figure \ref{fig:body_representation}a) in physical reasoning. Developmental studies also show that infants first carve the visual field into cohesive but rough volumetric entities with approximate spatial extent, before they infer contact relations, support, or motion trajectories~\cite{spelke1990principles, baillargeon2004infant}. This coarse-object representation has proven useful in machine learning as well, as the basis for computational studies that built models of human core knowledge \cite{smith2019modeling}. \\

\noindent While psychophysical studies \cite{li2023approximate} suggest that humans rely on approximate internal representations of object bodies, the nature and structure of these representations remain largely unknown. Behavioral methods can only offer indirect, low-resolution glimpses into these internal encodings, and cannot reveal their geometric or computational form. Humans construct their understanding of objects through a largely bottom-up visual process—segmenting, grouping, and localizing entities before reasoning about their dynamics and causal relations \citep{spelke1990principles, baillargeon2004infant}. Artificial segmentation models, which are trained to perform a similar decomposition of scenes into discrete entities, therefore offer a natural computational proxy for exploring how such representations might form in the human brain \cite{Gizdov_2025_CVPR}. If modern vision models trained for segmentation or prediction share representational structure with humans (which is \textit{not} clear a priori), they could help reveal the hidden organization of object representations that support intuitive physics and action. \\

\noindent Given this potential, a central question emerges: do the latent object representations that arise in vision models resemble the coarse body representations humans rely on for physical reasoning (see Figure \ref{fig:body_representation})? In nearly all segmentation models and datasets where an explicit teaching signal, reward, or loss function is used, the gold standard for accuracy is ground-truth segmentation or pixel-perfect human annotation. Such fine-grained segmentation may diverge from the approximate object representations that underpin human physical intuition. This misalignment is both inefficient and risky: fine-grain representations waste computation and may lead agents to mispredict human behavior, since people act based on intuitive, coarse-grain physics rather than exact geometric detail \cite{li2023approximate}. \textit{Unlike} in the study of language models, there has been little exploration of the alignment between the body representations humans use when reasoning about physics and the object representations artificial intelligence (AI) models use for vision, which is often a first bottom-up step for physical reasoning systems.

\noindent \textbf{Contributions.} In this work, we make the following contributions:

\begin{enumerate}
    \item \textbf{Object representations in vision models are similar to the body representations in humans}.
          We show that artificial segmentation networks form coarse body representations similar to those discovered in humans, particularly in the context of intuitive physical reasoning. To compare the two, we propose a framework that adapts a psychophysical experiment given to 50 human participants to vision models (Section \textit{'Human-like body representations in vision models'}).
    \item \textbf{Coarse body representations emerge naturally as a consequence of small network size}.
          We test 6 image segmentation architectures from the same family of models, varying in size, and discover that human-like coarse representations emerge as a consequence of small network size and limited training compute. We hypothesize that the resource-constrained nature of the human brain similarly favors efficient, coarse-grain representations that balance predictive power with computational efficiency (Section \textit{'Human-like representations as a consequence of resource constraints'}).
\end{enumerate}

\noindent This work provides a comparison of object representations in humans and vision models. Such alignment is important both for safety and interpretability in human–robot interaction, and for using machine models as computational probes of the brain’s internal representations, offering a scalable route to understanding how humans encode and reason about the physical world.

\section{Related Work}

People can reason efficiently about the physical dynamics of everyday objects, though they are also prone to systematic biases and errors under certain conditions \cite{kubricht2017intuitive}. This `intuitive physics' develops early, has a dedicated neural architecture, and is likely shared with non-human animals \cite{fischer2016functional, spelke2007core, spelke2022babies}. There are ongoing debates about the specific format of the representations and computations that support intuitive physics in humans, and proposals over the years have included (among others) first order logic, pre-Newtonian intuitive theories, heuristics and biases, and qualitative reasoning \cite{hartshorne2025insights}.

\noindent One prominent proposal for the computation that underlies human intuitive physics posits that people can carry out an internal mental simulation, similar to the computations that support engineered physical engines and game engines \cite{battaglia2013simulation, ullman2017mind}. This `mental game engine' proposal has been applied to people's reasoning about collisions, liquids, rigid- and soft-body motion, physical prediction, counterfactual and causal reasoning, and more \cite{smith2024probabilistic}. While research into mental simulation is ongoing, even if such a mental simulation exists, it cannot be perfectly accurate. Engineered simulations make heavy use of various approximations and workarounds, and it is likely that people's mental simulation uses approximations as well \citep{ullman2017mind, balaban2025physics, wang2025resource, bass2021partial}. \\

\noindent One major approximation in simulated environments is the use of simplified objects for the purposes of physical tracking. To clarify by example: While an advanced game may use fine-grain meshes to graphically display high-resolution images, it will often use only rough bodies for the purposes of collision detection. In line with this, there are theoretical reasons \cite{ullman2017mind} and recent empirical evidence \cite{li2023approximate} to suggest that people also make use of approximate object representations in physical reasoning.

\noindent Paralleling the interest in cognitive science, researchers in machine learning have studied the possibility of endowing machines with a sense of intuitive physics. Many datasets and challenges exist in this domain \citep[e.g.][]{bear2021physion,bakhtin2019phyre,yi2019clevrer,riochet2021intphys}, and several different frameworks have been proposed, ranging from those that emphasize built-in structure \cite{smith2019modeling} to those that emphasize learned representations \cite{piloto2022intuitive,garrido2025intuitive}, with various hybrid proposals in between \cite{duan2022survey}.\\

\noindent Despite this interest in machine learning, there hasn't been direct investigation (to the best of our knowledge) of whether the approximate object representations learned by vision models match the approximations used by humans in the context of physical reasoning. \citet{smith2019modeling} explicitly uses approximate bodies and credits them with successful generalization, but does not compare these to people. \citet{li2023approximate} used a model based on $\alpha$-shapes to examine people's object approximations, but does not explicitly endorse this as a cognitive model, but as a way of teasing apart different degrees of approximation.

\noindent Aside from offering glimpses into potential parallels between human and model representations, our work takes a more detailed view of how such representations evolve during training. Rather than focusing on global alignment metrics, we investigate the micro-level dynamics of segmentation learning—how sensitivity to different geometric structures, particularly concavities, changes as models grow or train longer. This approach complements broader forecasting efforts that aggregate model performance as a function of scale or time \citep{dl_is_predictable, time_to_fifty_percent, sevilla_forecast}, by revealing the finer representational shifts that underlie those macro-level trends

\section{Methods}
\label{sec:methods}

\subsection{Datasets and training paradigm}
\label{sec:data}
\begin{figure}[t]
  \centering
  \begin{minipage}[b]{0.58\linewidth}
    \centering
    \includegraphics[width=\linewidth]{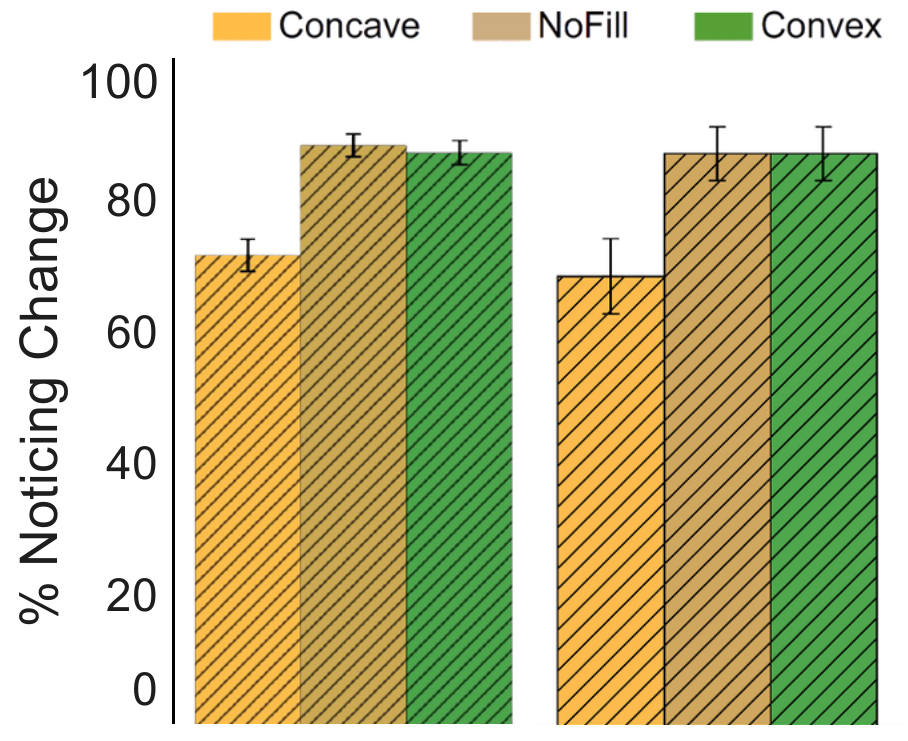}
    \vspace{2pt}
    {\footnotesize (a) Humans vs DETR vision model.}
  \end{minipage}\hfill
  \begin{minipage}[b]{0.42\linewidth}
    \centering
    \includegraphics[width=\linewidth]{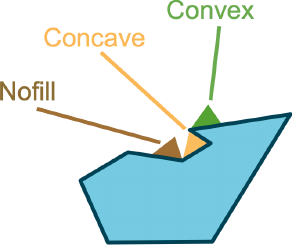}
    \vspace{2pt}
    {\footnotesize (b)}
  \end{minipage}
  \caption{(a) Change detection experiment: Humans (left) vs.\ Model (right). (b) A small local piece added to one of three locations: Nofill, Concave, and Convex body parts.}
  \label{fig:exp3b_and_categories}
\end{figure}

\noindent \textbf{Human–tested dataset.} Here we revisit prior work by \citet{li2023approximate} that this study builds upon. Each trial performed on a human subject (50 total participants) consists of a pair of black background RGB images of a polygon (see Figure \ref{fig:body_representation}): \emph{before} ($I_{\text{init}}$) and \emph{after} ($I_{\text{out}}$). $I_{\text{out}}$ is the same image as $I_{\text{init}}$, only with a small segment added (or not) at one of three locations: \textsc{concave}, \textsc{Nofill}, or \textsc{convex} (see Figure \ref{fig:exp3b_and_categories}b). Participants are shown $I_{\text{init}}$ for 1s, then a blank screen for 2s, then $I_{\text{out}}$ for another 1s. The participants are then asked to tell whether the polygon changed, and their accuracy is measured across the three conditions listed. We refer the reader to \citet{li2023approximate}, Experiment 3b, for a more detailed description of how the human data was collected. In this section, we aim to establish a pipeline for adapting pure segmentation models to the same experimental task. \\

\noindent \textbf{Model training dataset.}
To compare model and human representations, we used the experimental stimuli from \citet{li2023approximate} for evaluation. Because psychophysical datasets are too small to train deep networks, we generated a larger synthetic dataset designed to approximate their geometric and visual statistics. The goal was not to reproduce the exact human stimuli, but to expose models to similar shape statistics and scene composition. Each training image contains a single uniformly colored polygon on a black background, making our model more accustomed to the data shown to humans. The dataset will be released publicly upon paper acceptance. \\

\noindent \textbf{Synthetic dataset generation.}
We developed a procedural polygon generator to produce geometrically diverse yet controlled stimuli. Each polygon is created by sampling:
(1) a vertex count uniformly from 5–12;
(2) a number of concavities (0–3); and (3) irregularity and spikiness parameters controlling local curvature and edge variance. Polygons are rendered on black backgrounds using one of 24 bright colors sampled from a palette matched to the luminance distribution of the experimental stimuli. The generator thus produces a broad range of shapes that maintain the key structural properties of the human-tested stimuli while preventing any overlap between training and evaluation data. It will be publicly available upon paper acceptance.

\subsection{Models and fine-tuning}
We use publicly available SegFormer models that were pre-trained on the ADE20K dataset \cite{zhou2017scene,xie2021segformer}: a hierarchical transformer encoder with a lightweight MLP decoder. We tested six sizes (B0--B5) with parameter counts $\sim$3.8M (B0) to $\sim$84.7M (B5), covering over an order of magnitude in capacity. We then fine-tune each variant on our synthetic polygon dataset using custom training settings: AdamW optimizer with learning rate $5 \times 10^{-5}$, cosine learning rate schedule with warmup, batch size of 4, weight decay of 0.01, and 15 epochs. The training was performed on NVIDIA A100 GPUs with mixed precision (bfloat16) and TF32 optimizations enabled. The models are trained with a combination of cross-entropy and Dice loss to segment the binary segmentation task. Additionally, we tested a publicly available version of the DETR model \cite{carion2020end} for a total of 7 architectures.

\begin{figure*}[t]
  \centering
  \includegraphics[width=\textwidth]{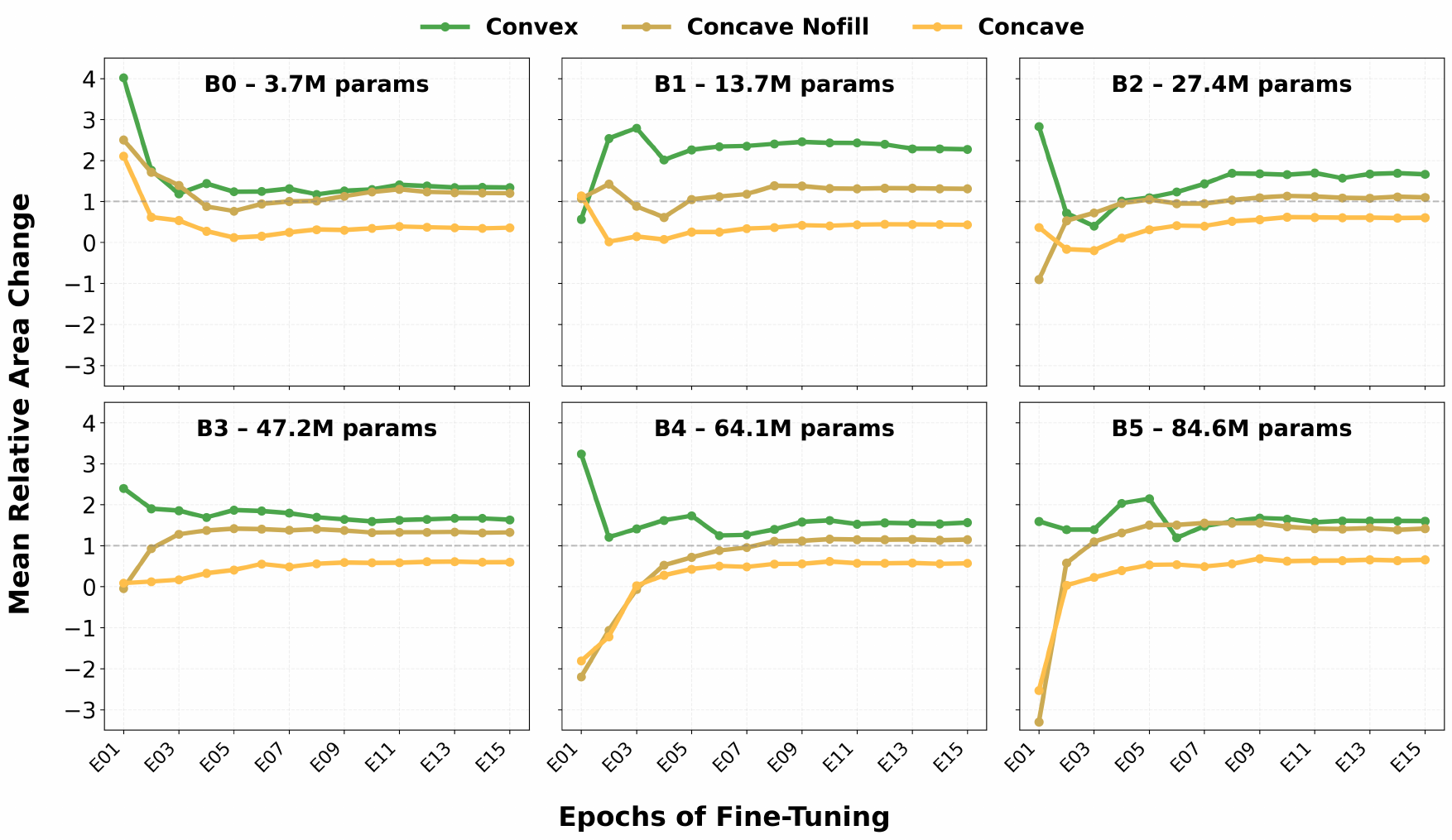}
  \caption{Mean $\mathrm{RAC}_{\text{seg}}$ during fine-tuning per category across models.}
  \label{fig:area_change_epochs}
\end{figure*}

\subsection{From model masks to change detection}
\label{sec:eval}

\textbf{Mask extraction.}
For each image $I\in\{I_{\text{init}},I_{\text{out}}\}$ the model outputs logits $\mathbf{L}(I)$. 
We apply softmax and argmax to get class predictions, then extract the foreground class (1: object class; 0: background class):
\[
\mathbf{M}^{(\text{pred})}(I) = \mathbf{1}\!\left[\text{argmax}(\text{softmax}(\mathbf{L}(I))) = 1\right]
\]
The object area is computed as the sum of object class pixels:
\[
A_{\text{init}} = \sum_{i,j} \mathbf{M}^{(\text{pred})}(I_{\text{init}})_{i,j},\quad
A_{\text{out}} = \sum_{i,j} \mathbf{M}^{(\text{pred})}(I_{\text{out}})_{i,j}
\]

\noindent \textbf{Relative Area Change (segment–normalized).}
Let $A_{\text{seg}}^{(\text{gt})}$ be the ground–truth pixel area of the \emph{edited segment} (the small local piece added/removed between $I_{\text{init}}$ and $I_{\text{out}}$, see Figure \ref{fig:exp3b_and_categories}). We define the \textit{Relative Area Change (RAC)}:
\[
\mathrm{RAC}_{\text{seg}}=\frac{A_{\text{out}}-A_{\text{init}}}{A_{\text{seg}}^{(\text{gt})}}.
\]
$\mathrm{RAC}_{\text{seg}}>0$ for additions (mask grows), $\mathrm{RAC}_{\text{seg}}<0$ for removals (mask shrinks), and $\mathrm{RAC}_{\text{seg}}$ measures how strongly the mask responds \emph{locally} at the manipulated region. A simple way to think about this is the local \textit{resolution} or \textit{sensitivity} of a model. A model that perfectly segments the ground truth will always have $\mathrm{RAC}_{\text{seg}}=1$, while a model incapable of perceiving a given local change will always have $\mathrm{RAC}_{\text{seg}} \approx 0$.
\\

\noindent \textbf{Change/no–change decision.}
We label a pair \{$I_{\text{init}}$, $I_{\text{out}}$\} \textsc{detected} iff $\mathrm{RAC}_{\text{seg}}>\tau$, with
$\tau\in\{0.1,0.2,0.3,0.4,0.5,...,20\}\%$. Each $\tau$ allows us to derive a change/no-change decision for each condition: \textsc{concave}, \textsc{Nofill}, \textsc{convex}.

\begin{figure*}[t]
  \centering
  \includegraphics[width=\textwidth]{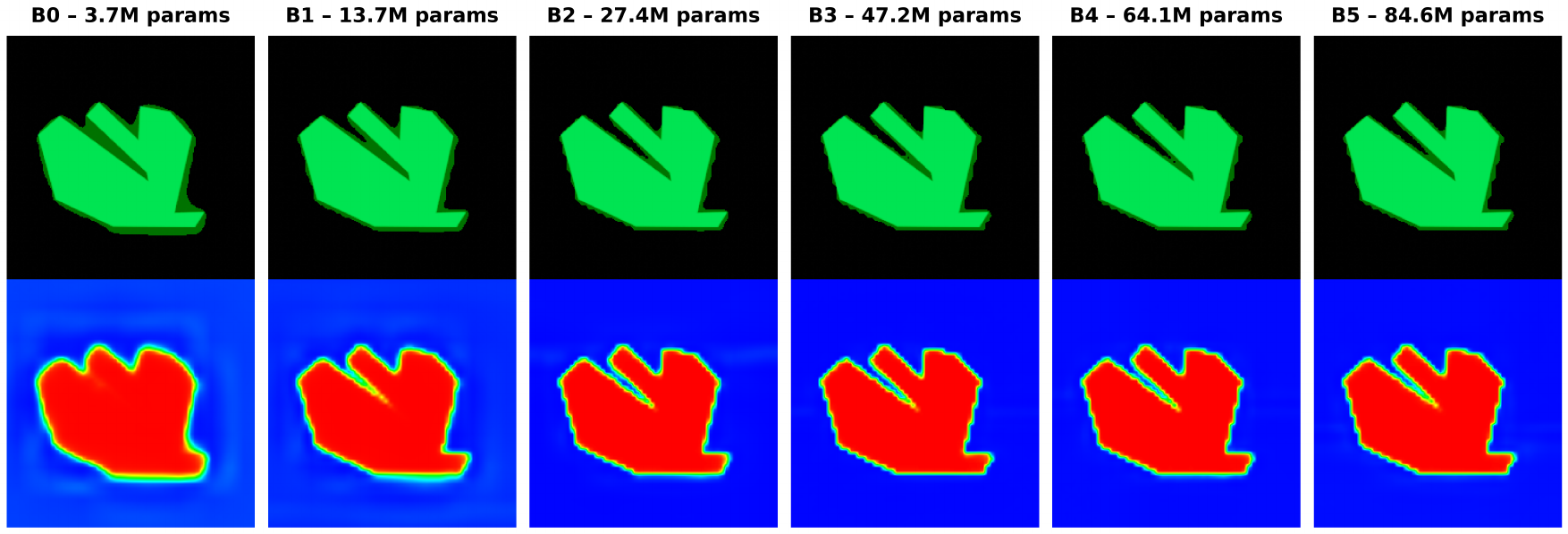}
  \caption{Mask overlays (first row) and probability heatmaps after 10 epochs of training across models.}
  \label{fig:collage_overlays_probs}
\end{figure*}

\section{Human-like representations in vision models}
\label{sec:human_like_representations}
We begin this section by stating an important takeaway from the above work by \citet{li2023approx}; people's representation of concave body parts appears more coarse than their representation of convex body parts. That is, people tend to "fill in" or "diffuse" concavities in the context of physical reasoning, which doesn't appear to be the case with convex body parts (see Figure \ref{fig:body_representation}). In the context of the change/no-change experiment discussed above, this is particularly evident in Figure \ref{fig:exp3b_and_categories}a (left). We see that the same holds for image segmentation models following the pipeline described above in Figure \ref{fig:exp3b_and_categories}a (right). This result is dependent on $\tau=1\%$, which is the optimal threshold value that minimizes the RMSE with human data, but given that we apply \textit{the same} $\tau$ to concave, nofill, and convex changes, we are simply showing that the model perceives changes in concavities as smaller, analogous to the more coarse representations formed by humans. \\

\noindent \textbf{Probability maps.} We refer the reader to Figure~\ref{fig:collage_overlays_probs}. On the \emph{top row}, segmented masks show clear outward diffusion around concavities, while corners and convex edges remain stable. The same pattern appears in the \emph{bottom row} of probability maps, where activations spread beyond concave boundaries. This consistent “filling-in” effect suggests that models, like humans, simplify concave regions into smoother, coarser body representations. \\

\noindent \textbf{Training dynamics.} As an additional test, we measured the \textit{average} $\mathrm{RAC}_{\text{seg}}$ on the test set for each change type throughout fine-tuning (Figure~\ref{fig:area_change_epochs}). Concave changes consistently remain at lower $\mathrm{RAC}_{\text{seg}}$ values (y-axis), even with extended training, indicating that models remain less responsive to changes inside concavities. We hypothesize that this reflects a representational bias toward compact, convex-like encodings. Representing objects through their convex hulls—or approximations close to it—requires fewer vertices and less spatial detail \cite{Duan_2015_CVPR}, reducing both memory load and computational cost. Moreover, such coarse representations generalize more effectively across object categories by capturing global shape structure rather than local irregularities. This consistent pattern between humans and vision models may hint at the underlying mechanisms and specific geometric forms that approximate body representations take in the human brain.

\subsection{Human-like representations as a consequence of resource constraints}
\noindent \textbf{Probability maps.} We refer the reader to Figure~\ref{fig:collage_overlays_probs}. Notably, the “filling-in’’ effect discussed above is much stronger in smaller models, whereas larger models preserve sharper boundaries and show reduced diffusion in concavities.
\\

\noindent \textbf{Training dynamics.} Similarly, we measured the \textit{average} $\mathrm{RAC}_{\text{seg}}$ for each change type throughout the fine-tuning (Figure~\ref{fig:area_change_epochs}). The gap between concave and convex changes diminishes both with increased training compute and with larger model size. As models grow and train longer, they become more sensitive to local geometric variations and less reliant on coarse, convex approximations. This pattern supports our hypothesis that human-like body representations emerge partly as a consequence of resource constraints: smaller or less-trained models favor compact, efficient representations that smooth concavities, while larger, more capable systems can afford finer geometric detail. \\

\noindent Together, these findings suggest that human-like coarse body representations may reflect an efficient encoding strategy that emerges naturally when computational or biological systems must balance representational detail with resource efficiency.

\section{Discussion}
People reason about the physical world using approximate rather than exact representations of objects. These coarse body representations are efficient for predicting how objects move or interact, without encoding every geometric detail. In this work, we asked whether such approximations naturally emerge in vision models trained for segmentation.

\noindent Across experiments, we found consistent parallels between human and model behavior. Both humans and models tend to simplify concavities, effectively “filling in’’ missing regions and producing smoother object representations. Quantitatively, models show lower sensitivity to local changes inside concavities, similar to the coarse representations observed in people's reasoning about intuitive physics.

\noindent Importantly, these effects vary with model capacity and training compute. Smaller networks and shorter training produce stronger concavity-smoothing effects, while larger or more extensively trained models develop sharper, more fine-grained boundaries. The same effect is \textit{not} observed for convex object parts, which are more stable and detailed. This pattern supports our hypothesis that human-like coarse body representations can emerge as an efficient solution under resource constraints: when capacity is limited, both biological and artificial systems favor compact, convex-like encodings that balance accuracy with computational efficiency.

\noindent Understanding this tradeoff may help clarify how efficient object representations arise in both minds and machines. While precise segmentations are essential for many applications, there is also value in representations that are “not too fine, not too coarse” — the kind that support intuitive, generalizable physical reasoning. This balance, much like in the tale of Goldilocks, may reflect the sweet spot of perception. \\

\bibliography{aaai2026}

@article{spelke1990principles,
  author  = {Spelke, Elizabeth S.},
  title   = {Principles of object perception},
  journal = {Cognitive Science},
  volume  = {14},
  number  = {1},
  pages   = {29--56},
  year    = {1990}
}

@book{marr2010vision,
  title={Vision: A computational investigation into the human representation and processing of visual information},
  author={Marr, David},
  year={2010},
  publisher={MIT press}
}

@article{baillargeon2004infant,
  author  = {Baillargeon, Ren{\'e}e},
  title   = {Infants’ reasoning about hidden objects: Evidence for event-general and event-specific expectations},
  journal = {Developmental Science},
  volume  = {7},
  number  = {4},
  pages   = {391--414},
  year    = {2004},
  doi     = {10.1111/j.1467-7687.2004.00357.x}
}

@article{battaglia2013simulation,
  author  = {Battaglia, Peter W. and Hamrick, Jessica B. and Tenenbaum, Joshua B.},
  title   = {Simulation as an engine of physical scene understanding},
  journal = {Proceedings of the National Academy of Sciences},
  volume  = {110},
  number  = {45},
  pages   = {18327--18332},
  year    = {2013},
  doi     = {10.1073/pnas.1306572110}
}

@article{smith2019modeling,
  title={Modeling expectation violation in intuitive physics with coarse probabilistic object representations},
  author={Smith, Kevin and Mei, Lingjie and Yao, Shunyu and Wu, Jiajun and Spelke, Elizabeth and Tenenbaum, Josh and Ullman, Tomer},
  journal={Advances in neural information processing systems},
  volume={32},
  year={2019}
}

@article{freyd1987dynamic,
  title={Dynamic mental representations.},
  author={Freyd, Jennifer J},
  journal={Psychological review},
  volume={94},
  number={4},
  pages={427},
  year={1987},
  publisher={American Psychological Association}
}

@article{balaban2025physics,
  title={Physics versus graphics as an organizing dichotomy in cognition},
  author={Balaban, Halely and Ullman, Tomer D},
  journal={Trends in Cognitive Sciences},
  year={2025},
  publisher={Elsevier}
}

@article{li2023approximate,
  author  = {Li, Yichen and Wang, YingQiao and Boger, Tal and Smith, Kevin A. and Gershman, Samuel J. and Ullman, Tomer D.},
  title   = {An approximate representation of objects underlies physical reasoning},
  journal = {Journal of Experimental Psychology: General},
  volume  = {152},
  number  = {11},
  pages   = {3074--3086},
  year    = {2023},
  doi     = {10.1037/xge0001439}
}

@article{kubricht2017intuitive,
  title={Intuitive physics: Current research and controversies},
  author={Kubricht, James R and Holyoak, Keith J and Lu, Hongjing},
  journal={Trends in cognitive sciences},
  volume={21},
  number={10},
  pages={749--759},
  year={2017},
  publisher={Elsevier}
}

@article{fischer2016functional,
  title={Functional neuroanatomy of intuitive physical inference},
  author={Fischer, Jason and Mikhael, John G and Tenenbaum, Joshua B and Kanwisher, Nancy},
  journal={Proceedings of the national academy of sciences},
  volume={113},
  number={34},
  pages={E5072--E5081},
  year={2016},
  publisher={National Academy of Sciences}
}

@book{spelke2022babies,
  title={What babies know: Core knowledge and composition volume 1},
  author={Spelke, Elizabeth},
  volume={1},
  year={2022},
  publisher={Oxford University Press}
}

@article{hartshorne2025insights,
  title={Insights into cognitive mechanics from education, developmental psychology and cognitive science},
  author={Hartshorne, Joshua K and Jing, Mengguo},
  journal={Nature Reviews Psychology},
  pages={1--15},
  year={2025},
  publisher={Nature Publishing Group US New York}
}

@article{ullman2017mind,
  title={Mind games: Game engines as an architecture for intuitive physics},
  author={Ullman, Tomer D and Spelke, Elizabeth and Battaglia, Peter and Tenenbaum, Joshua B},
  journal={Trends in cognitive sciences},
  volume={21},
  number={9},
  pages={649--665},
  year={2017},
  publisher={Elsevier}
}

@article{spelke2007core,
  title={Core knowledge},
  author={Spelke, Elizabeth S and Kinzler, Katherine D},
  journal={Developmental science},
  volume={10},
  number={1},
  pages={89--96},
  year={2007},
  publisher={Wiley Online Library}
}

@article{duan2022survey,
  title={A survey on machine learning approaches for modelling intuitive physics},
  author={Duan, Jiafei and Dasgupta, Arijit and Fischer, Jason and Tan, Cheston},
  journal={arXiv preprint arXiv:2202.06481},
  year={2022}
}

@article{garrido2025intuitive,
  title={Intuitive physics understanding emerges from self-supervised pretraining on natural videos},
  author={Garrido, Quentin and Ballas, Nicolas and Assran, Mahmoud and Bardes, Adrien and Najman, Laurent and Rabbat, Michael and Dupoux, Emmanuel and LeCun, Yann},
  journal={arXiv preprint arXiv:2502.11831},
  year={2025}
}

@article{piloto2022intuitive,
  title={Intuitive physics learning in a deep-learning model inspired by developmental psychology},
  author={Piloto, Luis S and Weinstein, Ari and Battaglia, Peter and Botvinick, Matthew},
  journal={Nature human behaviour},
  volume={6},
  number={9},
  pages={1257--1267},
  year={2022},
  publisher={Nature Publishing Group UK London}
}

@article{yi2019clevrer,
  title={Clevrer: Collision events for video representation and reasoning},
  author={Yi, Kexin and Gan, Chuang and Li, Yunzhu and Kohli, Pushmeet and Wu, Jiajun and Torralba, Antonio and Tenenbaum, Joshua B},
  journal={arXiv preprint arXiv:1910.01442},
  year={2019}
}

@article{riochet2021intphys,
  title={Intphys 2019: A benchmark for visual intuitive physics understanding},
  author={Riochet, Ronan and Castro, Mario Ynocente and Bernard, Mathieu and Lerer, Adam and Fergus, Rob and Izard, V{\'e}ronique and Dupoux, Emmanuel},
  journal={IEEE Transactions on Pattern Analysis and Machine Intelligence},
  volume={44},
  number={9},
  pages={5016--5025},
  year={2021},
  publisher={IEEE}
}

@article{bear2021physion,
  title={Physion: Evaluating physical prediction from vision in humans and machines},
  author={Bear, Daniel M and Wang, Elias and Mrowca, Damian and Binder, Felix J and Tung, Hsiao-Yu Fish and Pramod, RT and Holdaway, Cameron and Tao, Sirui and Smith, Kevin and Sun, Fan-Yun and others},
  journal={arXiv preprint arXiv:2106.08261},
  year={2021}
}

@article{bakhtin2019phyre,
  title={Phyre: A new benchmark for physical reasoning},
  author={Bakhtin, Anton and van der Maaten, Laurens and Johnson, Justin and Gustafson, Laura and Girshick, Ross},
  journal={Advances in Neural Information Processing Systems},
  volume={32},
  year={2019}
}

@article{wang2025resource,
  title={Resource bounds on mental simulations: Evidence from a liquid-reasoning task.},
  author={Wang, YingQiao and Ullman, Tomer D},
  journal={Journal of Experimental Psychology: General},
  year={2025},
  publisher={American Psychological Association}
}

@article{bass2021partial,
  title={Partial mental simulation explains fallacies in physical reasoning},
  author={Bass, Ilona and Smith, Kevin A and Bonawitz, Elizabeth and Ullman, Tomer D},
  journal={Cognitive Neuropsychology},
  volume={38},
  number={7-8},
  pages={413--424},
  year={2021},
  publisher={Taylor \& Francis}
}

@article{smith2024probabilistic,
  title={Probabilistic models of physical reasoning},
  author={Smith, Kevin A and Hamrick, Jessica B and Sanborn, Adam N and Battaglia, Peter W and Gerstenberg, Tobias and Ullman, Tomer D and Tenenbaum, Joshua B},
  journal={Reverse engineering the mind: Probabilistic models of cognition},
  year={2024}
}

@inproceedings{xie2021segformer,
  title        = {SegFormer: Simple and Efficient Design for Semantic Segmentation with Transformers},
  author       = {Xie, Enze and Wang, Wenhai and Yu, Zhiding and Anandkumar, Anima and Alvarez, Jose M. and Luo, Ping},
  booktitle    = {Advances in Neural Information Processing Systems (NeurIPS)},
  year         = {2021},
  url          = {https://arxiv.org/abs/2105.15203}
}

@inproceedings{carion2020end,
  title={End-to-end object detection with transformers},
  author={Carion, Nicolas and Massa, Francisco and Synnaeve, Gabriel and Usunier, Nicolas and Kirillov, Alexander and Zagoruyko, Sergey},
  booktitle={European conference on computer vision},
  pages={213--229},
  year={2020}
}

@InProceedings{Gizdov_2025_CVPR,
    author    = {Gizdov, Andrey and Ullman, Shimon and Harari, Daniel},
    title     = {Seeing More with Less: Human-like Representations in Vision Models},
    shortauthor = {Gizdov},
    booktitle = {Proceedings of the IEEE/CVF Conference on Computer Vision and Pattern Recognition (CVPR)},
    month     = {June},
    year      = {2025},
    pages     = {4408-4417}
}

@misc{dl_is_predictable,
      title={Deep Learning Scaling is Predictable, Empirically}, 
      author={Joel Hestness and Sharan Narang and Newsha Ardalani and Gregory Diamos and Heewoo Jun and Hassan Kianinejad and Md. Mostofa Ali Patwary and Yang Yang and Yanqi Zhou},
      year={2017},
      url={https://arxiv.org/abs/1712.00409}, 
}

@misc{time_to_fifty_percent,
      title={Measuring AI Ability to Complete Long Tasks}, 
      author={Thomas Kwa and Ben West and Joel Becker and Amy Deng and Katharyn Garcia and Max Hasin and Sami Jawhar and Megan Kinniment and Nate Rush and Sydney Von Arx and Ryan Bloom and Thomas Broadley and Haoxing Du and Brian Goodrich and Nikola Jurkovic and Luke Harold Miles and Seraphina Nix and Tao Lin and Neev Parikh and David Rein and Lucas Jun Koba Sato and Hjalmar Wijk and Daniel M. Ziegler and Elizabeth Barnes and Lawrence Chan},
      year={2025},
      url={https://arxiv.org/abs/2503.14499}, 
}

@inproceedings{sevilla_forecast,
   title={Compute Trends Across Three Eras of Machine Learning},
   booktitle={2022 International Joint Conference on Neural Networks (IJCNN)},
   author={Sevilla, Jaime and Heim, Lennart and Ho, Anson and Besiroglu, Tamay and Hobbhahn, Marius and Villalobos, Pablo},
   year={2022},}

@InProceedings{Duan_2015_CVPR,
author = {Duan, Liuyun and Lafarge, Florent},
title = {Image Partitioning Into Convex Polygons},
booktitle = {Proceedings of the IEEE Conference on Computer Vision and Pattern Recognition (CVPR)},
month = {June},
year = {2015}
}

@article{li2023approx,
  title        = {An Approximate Representation of Objects Underlies Physical Reasoning},
  author       = {Li, Yichen and Wang, YingQiao and Boger, Tal and Smith, Kevin A. and Gershman, Samuel J. and Ullman, Tomer D.},
  journal      = {Journal of Experimental Psychology: General},
  year         = {2023},
  doi          = {10.1037/xge0001439},
  url          = {https://talboger.github.io/files/Li_etal_JEPG_2023.pdf}
}

@inproceedings{zhou2017scene,
  author    = {Bolei Zhou and Hang Zhao and Xavier Puig and Sanja Fidler and Adela Barriuso and Antonio Torralba},
  title     = {Scene Parsing through ADE20K Dataset},
  booktitle = {Proceedings of the IEEE Conference on Computer Vision and Pattern Recognition (CVPR)},
  year      = {2017},
  organization = {IEEE}
}

\end{document}